\documentclass[10pt,twocolumn,letterpaper]{article}
\usepackage{cvpr}
\definecolor{cvprblue}{rgb}{0.21,0.49,0.74}
\usepackage[pagebackref,breaklinks,colorlinks,allcolors=cvprblue]{hyperref}
\usepackage{multirow}
\usepackage{times}
\usepackage{graphicx}
\usepackage{amsmath,amssymb}
\usepackage{booktabs}
\usepackage{enumitem}
\usepackage{placeins}
\usepackage{float}
\usepackage{ragged2e}
\usepackage[labelfont=bf]{caption}
\RequirePackage[numbers,sort&compress]{natbib}
\usepackage{xcolor}

\definecolor{darkred}{RGB}{139,0,0}

\def\paperID{17381}
\def\confName{CVPR}
\def\confYear{2026}

\title{\vspace{-2.5mm}Diffusion Representations for Fine-Grained Image Classification:\\ A Marine Plankton Case Study\vspace{-2.5mm}}

\author{%
  Aleix Nieto Juscafresa\thanks{Corresponding author.}\,\;\thanks{Equal contribution.}
  \qquad
  Álvaro Mazcuñán Herreros\footnotemark[2] 
  \qquad
  Josephine Sullivan \\
  KTH Royal Institute of Technology, Stockholm, Sweden \\
  \texttt{\{aleixnj, alvaromh, sullivan\}@kth.se}
}

\begin{document}
\maketitle

\begin{abstract}
Diffusion models have emerged as state-of-the-art generative methods for image synthesis, yet their potential as general-purpose feature encoders remains underexplored. Trained for denoising and generation without labels, they can be interpreted as self-supervised learners that capture both low- and high-level structure. We show that a frozen diffusion backbone enables strong fine-grained recognition by probing intermediate denoising features across layers and timesteps and training a linear classifier for each pair. We evaluate this in a real-world plankton-monitoring setting with practical impact, using controlled and comparable training setups against established supervised and self-supervised baselines. Frozen diffusion features are competitive with supervised baselines and outperform other self-supervised methods in both balanced and naturally long-tailed settings. Out-of-distribution evaluations on temporally and geographically shifted plankton datasets further show that frozen diffusion features maintain strong accuracy and Macro F1 under substantial distribution shift.
\end{abstract}

\section{Introduction}
Diffusion models \cite{ho2020ddpm, pmlr-v37-sohl-dickstein15, song2020score} have become a leading approach for generative modeling, delivering remarkable results in image synthesis \cite{dhariwal2021diffusion, ho22classifier, fuest2024diffusion} and extending to other modalities such as audio \cite{kong2020diffwave, mittal2021symbolic}, reinforcement learning \cite{chi2025diffusion, hansen2023idql, zhu2023diffusion}, and computational biology  \cite{wu2022diffusion, gruver2023protein}. Through iterative denoising, diffusion models learn representations that capture global structure and fine-grained textures \cite{yue2024exploring}. In their label-free training approach, they provide off-the-shelf, self-supervised features that transfer beyond generation and perform strongly in recognition tasks \cite{fuest2024diffusion}. Furthermore, several works demonstrate that these features can remain discriminative even without task-specific tuning \cite{xiang2023denoising, kim2025revelio}. Based on these findings, recent effort has gone into identifying which layers and timesteps are most discriminative for classification and which representation details they capture, noting contrasts between fine-grained and more global features such as shape \cite{kim2025revelio, luo2023diffusion}. 

Building on this line of work, we extend the analysis by evaluating diffusion representations in a \textit{fine-grained} regime and by characterizing the link between generative quality and discriminative performance. In particular, we revisit layer and timestep selection under fine-grained recognition, quantify how fidelity affects classification accuracy, and diagnose when overfitting during diffusion-model training breaks this link. We therefore aim to answer two questions:
\begin{itemize}[itemsep=3pt]
\item \textit{How well do pretrained diffusion features support fine-grained recognition across balanced and imbalanced training regimes?}
\item \textit{How are generation quality and the diffusion objective, including its per-timestep loss distribution, related to the discriminative strength of the resulting representations?}
\end{itemize}

To ground these questions in a fine-grained domain, we focus on plankton imagery. It poses an extreme fine-grained challenge with severe class imbalance (a few abundant taxa and a long tail of rare species), low contrast, and subtle inter-class variation \cite{Eerola2024PlanktonSurvey}. 

Furthermore, beyond the technical challenge, \textit{advances in plankton recognition directly benefit marine ecosystem monitoring}. This is because plankton communities control key ocean processes and serve as sensitive markers of climate-driven change, including warming, acidification, and shifting nutrient regimes \cite{Falkowski1998, Beaugrand2002, Doney2009}. To track these shifts at scale, automated instruments now produce massive, low-contrast image streams that must be classified by taxon to support abundance estimates and trend analysis \cite{olson2007submersible, Orenstein2015WHOIPlankton}. Labels are scarce and the task is fine-grained: taxa may differ only by slight changes, small spine or flagellum differences, and faint internal banding or vacuoles \cite{Eerola2024PlanktonSurvey}. 

To address these challenges in low-contrast plankton imagery, we build on DDAE \cite{xiang2023denoising}, a denoising diffusion feature extractor that requires no fine-tuning, to identify which U-Net decoder activations yield the highest linear-probe accuracy. We sweep decoder layers and timesteps, keep features frozen, and evaluate with a linear probe.

We assess the layer-by-timestep readout across a curated class-balanced benchmark, a realistic long-tailed scenario with under-represented classes, and an out-of-distribution (OOD) test that transfers to a distinct plankton domain with unseen classes. In this out-of-distribution scenario, the source and target datasets differ in acquisition year, environmental and imaging conditions, and geographic location, so that the evaluation probes how robust the learned representations are. Across all three settings, our diffusion features outperform self-supervised baselines such as DINOv3 \cite{simeoni2025dinov3} under our evaluation protocol and are competitive with supervised baselines trained end-to-end, including Vision Transformers (ViTs) \cite{dosovitskiy2020image}, ResNets \cite{he2016deep}, and EfficientNet \cite{tan2019efficientnet}. By avoiding plankton-specific challenges \cite{Eerola2024PlanktonSurvey}, the study measures representation quality and provides practical guidance on where and when to select self-supervised features to support recognition effectively, including under realistic distribution shifts.

\vspace{0.75mm}
\hspace{-0.5cm} We summarize our main contributions as follows:
\begin{enumerate}
\item We formalize the general problem of selecting the optimal layer–timestep pair for feature selection, and, using a linear probe, we find a stable low-depth, moderate-noise region where diffusion features are most useful; we test this framework in fine-grained recognition.\vspace{1mm}
\item We evaluate curated, imbalanced, and cross-distribution plankton regimes, amortizing compute by reusing precomputed descriptors across related datasets, matching or surpassing strong supervised and self-supervised baselines.\vspace{1mm}
\item We link generation quality to recognition, and to the optimal layer–timestep choice, by analyzing how the training objective and its noise-level loss schedule shape learned features.
\end{enumerate}

\section{Related Work}
\paragraph{Fine-Grained Representations for Classification.}
Fine-grained classification hinges on capturing subtle, part-level distinctions, requiring representations that preserve local micro-patterns while integrating global context. For this task, attention mechanisms have proven especially effective: RA-CNN \cite{fu2017look} sharpens part localization; CBAM \cite{woo2018cbam} reweights spatial and channel cues; TransFG \cite{he2022transfg} uses token attention to surface discriminative parts; ViT \cite{dosovitskiy2020image} and CrossViT \cite{chen2021crossvit} capture long-range, multi-scale relations; bilinear pooling \cite{lin2015bilinear} isolates localized pairwise textures. In diffusion U-Nets, self-attention is applied at every denoising step, allowing distant regions to repeatedly exchange detail as noise is removed \cite{dhariwal2021diffusion, yang2023diffusion}. This iterative process refines part correspondences and preserves fine texture, yielding compact, part-aware features that preserve micro-patterns under noise and clutter, motivating our use of frozen diffusion features for recognition \cite{yang2023diffusion, mukhopadhyay2023diffusion, han2024feature}.

\paragraph{Self-Supervised Learners for Representation.}
Self-supervised encoders learn without labels by exploiting image structure. Discriminative methods align crops from the same image; DINOv3 adds a momentum teacher, multi-crop views, centering/sharpening, and prototype assignments to yield robust, semantically organized features. Reconstruction methods predict missing content; MAE \cite{he2022masked} masks a large patch fraction and reconstructs with a lightweight decoder, concentrating capacity in the encoder. These families differ in inductive bias: discriminative objectives promote invariance to viewpoint, color, and background, whereas reconstruction preserves fine detail and layout, which is useful for fine-grained and dense tasks.

\paragraph{Diffusion Model Representations.}
Diffusion models \cite{ho2020ddpm, song2020score} are trained for generation but are hypothesized to learn strong representations while denoising across many noise scales \cite{fuest2024diffusion}. Competitive classification can be achieved by using intermediate activations without modifying the model \cite{xiang2023denoising}. Extending that idea, \cite{mukhopadhyay2024text} identifies informative layers and noise levels and shows that aggregating across layers and timesteps improves linear separability. Low-dimensional latent spaces and structured noise schedules emerge as design factors for discriminative features \cite{chen2024deconstructing}. Motivated by these observations, we study diffusion-derived features for plankton imagery, a low-contrast and fine-grained domain where such features have not been systematically evaluated. We report linear probes to measure representation quality independently of head capacity, following standard self-supervised practice used by DDAE \cite{xiang2023denoising} or DINOv3 \cite{simeoni2025dinov3}. 

\paragraph{Generative Approaches for Plankton Classification.}
Generative approaches for plankton classification have focused on augmentation and unsupervised representation learning, yet recognition gains remain limited. Conditional GANs synthesize minority taxa and report small improvements over convolutional baselines \cite{Wang2017CGANPlankton, Liu2018GANCurriculum}. CycleGAN increases the presence of rare Imaging FlowCytobot (IFCB) \cite{torstensson2024smhiifcb} taxa and can raise average precision once synthetic images are added to training \cite{Li2021PlanktonDetection, Liu2018GANCurriculum}. On very small algae collections, DCGAN with transfer learning improves accuracy after enlarging the training set, although effects are inconsistent across settings \cite{Khan2022DCGANAlgae, Ali2022AlgaeWater, Radford2015DCGAN}. Beyond GANs, a VAE pipeline clusters plankton images by extracting DenseNet features without fine-tuning, learning a latent code with a VAE, and grouping the low-dimensional embeddings with fuzzy $k$-means, which reveals meaningful structure but does not establish clear benefits for supervised IFCB recognition \cite{alfano2022efficient}. Motivated by scarce labels and morphology-driven differences in plankton imagery, we evaluate diffusion-derived representations for recognition, a direction that, to our knowledge, has not been explored on IFCB data.

\section{Background}\label{background}

We provide a brief overview of the original diffusion model formulation \cite{ho2020ddpm}. Since then, numerous variants have been proposed across its modular components, training objectives, noise parameterizations, and sampling procedures. For a detailed overview of these design choices, see \cite{cao2024survey}.

\paragraph{Forward (noising) Process.}
Diffusion models \cite{ho2020ddpm,nichol2021improved,song2020score} define a series of Gaussian corruptions on data $x_0$. For timesteps $t=1,\dots,T$, the corruption is described by the transition probability
\begin{equation}\label{forward}
q(x_t \mid x_{t-1}) = \mathcal N\!\big(\sqrt{\alpha_t}\,x_{t-1},\; (1-\alpha_t)\,\mathbf I\big),
\end{equation}
where $\alpha_t \in (0,1)$ sets the noise level. Conditioned on $x_0$, the forward noising distribution at step $t$ is $
q(x_t \mid x_0)
=
\mathcal N\!\big(x_t;\, \sqrt{\bar\alpha_t}\,x_0, (1-\bar\alpha_t)\,\mathbf I\big),$
where $\bar\alpha_t = \prod_{s=1}^t \alpha_s$.
Equivalently, using the reparameterization trick, a noised version of $x_0$ at an arbitrary level $t$ can be obtained by sampling $\varepsilon \sim \mathcal N(0,\mathbf I)$ and taking
$
x_t = \sqrt{\bar\alpha_t}\,x_0 + \sqrt{1-\bar\alpha_t}\,\varepsilon.
$
When $T$ is large, $x_T \approx \mathcal N(0,\mathbf I)$. Common schedules choose $\alpha_t$ to vary smoothly (e.g., linear or cosine \cite{nichol2021improved}).

\paragraph{Reverse (denoising) Process.}
At generation time we draw $x_T \sim \mathcal N(0,\mathbf I)$ and iteratively denoise via $x_t \!\to\! x_{t-1}$. The one-step reverse law is the marginal over the unknown clean sample $x_0$,
\begin{equation}
q(x_{t-1}\mid x_t)
= \int q(x_{t-1}\mid x_t,x_0)\,q(x_0\mid x_t)\,dx_0.
\end{equation}
Here $q(x_{t-1}\mid x_t,x_0)$ is Gaussian because, under the linear-Gaussian forward process, $(x_{t-1},x_t)$ are affine functions of $x_0$ plus Gaussian noise and hence jointly Gaussian; conditioning on $x_0$ preserves Gaussianity. The factor $q(x_0\mid x_t)$ reflects uncertainty from the (unknown) data distribution, so the marginal above is a mixture and not tractable in closed form. 

\setlength{\tabcolsep}{0pt}
\renewcommand{\arraystretch}{0}
\begin{figure*}
\centering
\begin{tabular}{c@{\hspace{14mm}}c@{\hspace{9mm}}c@{\hspace{9mm}}c}
    $x_0$ & $t=25$ & $t=200$ & $t=600$ \vspace{2mm} \\
  \includegraphics[width=0.095\textwidth]{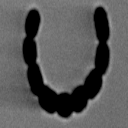} &
  \includegraphics[width=0.19\textwidth]{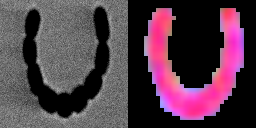} &
  \includegraphics[width=0.19\textwidth]{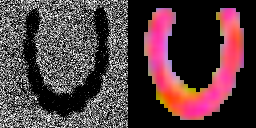}  &
  \includegraphics[width=0.19\textwidth]{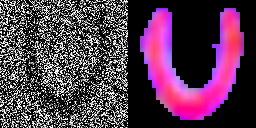}  \\
  
  \includegraphics[width=0.095\textwidth]{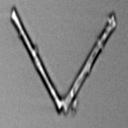} &
  \includegraphics[width=0.19\textwidth]{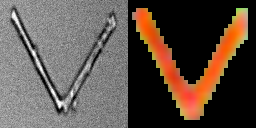} &
  \includegraphics[width=0.19\textwidth]{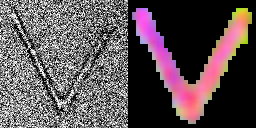}  &
  \includegraphics[width=0.19\textwidth]{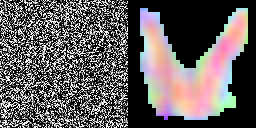}  \\
  
  \includegraphics[width=0.095\textwidth]{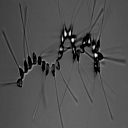} &
  \includegraphics[width=0.19\textwidth]{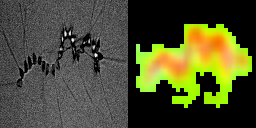} &
  \includegraphics[width=0.19\textwidth]{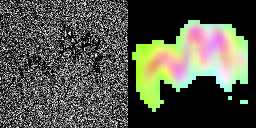}  &
  \includegraphics[width=0.19\textwidth]{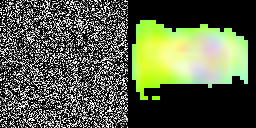}  \\
  
  \includegraphics[width=0.095\textwidth]{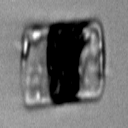} &
  \includegraphics[width=0.19\textwidth]{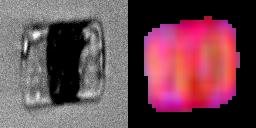} &
  \includegraphics[width=0.19\textwidth]{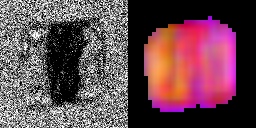}  &
  \includegraphics[width=0.19\textwidth]{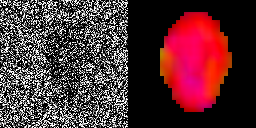}  \\
  
\end{tabular}
\caption{\textbf{Principal Component Analysis (PCA) visualization of diffusion features}. Using random samples, we compute PCA on the diffusion features and map the first three principal components to the RGB channels. For each sample, we show the clean target image and its noisy versions at timesteps $t \in \{25, 200, 600\}$ with the corresponding PCA colorization overlaid on the noisy image. Background regions are suppressed by masking pixels with low values on the first principal component. This visualization is inspired by \cite{oquab2024dinov2learningrobustvisual}.}
\label{fig:PCA_visualization}
\end{figure*}

Because the forward step in Equation \eqref{forward} is linear-Gaussian, small $\beta_t=1-\alpha_t$ makes each step an infinitesimal perturbation. In the continuous-time limit, time reversal yields a locally Gaussian reverse transition whose mean and covariance are determined by the score of the marginal $q_t$ \cite{pmlr-v37-sohl-dickstein15, song2020score}:
\vspace{-1mm}
\begin{equation}
m(x_t,t)\approx \frac{1}{\sqrt{1-\beta_t}}\Bigl(x_t+\beta_t \nabla_{x_t}\log q_t(x_t)\Bigr)
\end{equation}
\begin{equation}
\mathrm{Cov}\approx \beta_t \mathbf I,
\end{equation}
where “$\approx$” denotes a small-step surrogate for the discrete transition (it does \emph{not} claim the unconditional discrete transition is exactly Gaussian).

At the actual discrete step $t\!\to\!t-1$, conditioning on $x_0$ makes $(x_{t-1},x_t)$ jointly Gaussian because they are affine functions of $x_0$ and Gaussian noises; therefore the conditional reverse posterior is Gaussian with density
$
q(x_{t-1}\mid x_t,x_0)=\mathcal{N}\!\bigl(x_{t-1};\,\tilde{\mu}_t(x_t,x_0),\,\tilde{\beta}_t \mathbf{I}\bigr),
$
where
\begin{equation}
\tilde{\mu}_t(x_t,x_0)
= \frac{\sqrt{\alpha_t}\,(1-\bar{\alpha}_{t-1})}{1-\bar{\alpha}_t}\, x_t
+ \frac{\sqrt{\bar{\alpha}_{t-1}}\,\beta_t}{1-\bar{\alpha}_t}\, x_0,
\end{equation}

\begin{equation}    
\tilde{\beta}_t
= \frac{\beta_t\,(1-\bar{\alpha}_{t-1})}{1-\bar{\alpha}_t}.
\end{equation}
Thus the \emph{per-step} variance $\tilde\beta_t \mathbf I$ is exact and depends only on the schedule, not on $x_t$ or $x_0$; a Taylor expansion shows $\tilde\beta_t=\beta_t-\frac{\bar\alpha_{t-1}}{1-\bar\alpha_{t-1}}\beta_t^{\,2}+O(\beta_t^{\,3})=\beta_t+O(\beta_t^{\,2})$. In practice we keep this variance fixed and compute the mean by plugging in a learned estimate $\hat x_0(x_t,t)$ obtained next:
\begin{equation}\label{estimate}
p_\theta(x_{t-1}\mid x_t)
=\mathcal N\!\bigl(x_{t-1};\,\tilde{\mu}_t\bigl(x_t,\hat x_0(x_t,t)\bigr),\,\tilde\beta_t \mathbf I\bigr).
\end{equation}

\paragraph{Noise Prediction (training objective).}
We parameterize a \textit{network} $E_\theta(x_t,t)$ to predict the Gaussian noise used to form $x_t$ and train it with the standard MSE $\mathcal L_{\text{noise}}(\theta) = \mathbb E_{t,x_0,\varepsilon}\!\left[\|\varepsilon - E_\theta(x_t,t)\|_2^2\right]$.
From $E_\theta$ we recover an estimate of the clean sample $
\hat x_0(x_t,t)
=
x_t - \sqrt{1-\bar\alpha_t}\,E_\theta(x_t,t)/\sqrt{\bar\alpha_t}$,
which is then substituted into $\tilde{\mu}_t(x_t,x_0)$ to produce the mean of $p_\theta(x_{t-1}\mid x_t)$ used during sampling. Equivalently, we may write the sampling mean directly in terms of the predicted noise as
\begin{align}
\mu_\theta(x_t,t) &= \tilde{\mu}_t\!\bigl(x_t,\hat x_0(x_t,t)\bigr)\\
&=
\frac{1}{\sqrt{\alpha_t}}\!\left(x_t-\frac{\beta_t}{\sqrt{1-\bar\alpha_t}}\,E_\theta(x_t,t)\right)
\end{align}

\paragraph{Sampling (iterative process).} Starting from a pure noise sample $x_T \sim \mathcal N(0,\mathbf I)$, use the learned Gaussian $p_\theta(x_{t-1}\!\mid\!x_t) = \mathcal N\!\big(\mu_\theta(x_t,t), \tilde\beta_t\,\mathbf I\big)$ and update $x_{t-1} \leftarrow \mu_\theta(x_t,t) + \tilde\beta_t^{1/2}\,z$, $z\!\sim\!\mathcal N(0,\mathbf I)$. 

This is the original DDPM sampler \cite{ho2020ddpm}; many improved variants exist \cite{cao2024survey}.

\paragraph{Diffusion Feature Extraction.}
The most common way of obtaining features is by first corrupting an input image to a chosen noise level $t$ to form $x_t$ and forwarding $(x_t,t)$ through a pretrained U-Net denoiser. Feature maps are read at the outputs of selected decoder residual blocks, providing multi-scale descriptors without updating model weights \cite{baranchuk2021label, namekata2024emerdiff, tang2023emergent, zhang2023tale, zhang2024telling}. Typical diffusion timesteps used in prior work include  $t=45$~\cite{xiang2023denoising}, $t=50$~\cite{zhang2024telling} or $t=261$~\cite{tang2023emergent}. Complementary to this noised-input paradigm, CleanDIFT shows that a lightweight, unsupervised fine-tuning of the diffusion backbone enables the extraction of high-quality semantic features directly from clean images--surpassing noise-augmented and even ensemble-based extraction at a fraction of the computational cost \cite{stracke2025cleandiftdiffusionfeaturesnoise}.

\section{Methodology}
\label{sec:method}
We use a diffusion model whose denoiser is a U-Net \textit{backbone} \cite{ronneberger2015u} with an encoder-bottleneck-decoder topology and skip connections. The U-Net is trained on our dataset (no pretraining) following the \textit{noise prediction} objective \cite{ho2020ddpm}. For a variance schedule $\{\beta_t\}_{t=1}^T$ with $\alpha_t = 1 - \beta_t$ and \textit{cumulative signal coefficient }$\bar{\alpha}_t = \prod_{s=1}^t \alpha_s$, the forward noising process \cite{ho2020ddpm} is
$x_t = \sqrt{\bar{\alpha}_t}\, x_0 + \sqrt{1 - \bar{\alpha}_t}\, \varepsilon,\ \varepsilon \sim \mathcal{N}(0, \mathbf I)$, where $x_t$ is the noised $x_0\sim p_{\text{data}}$ and $t$ indexes the noise level. Let $E_\theta(x_t, t)$ denote the noise estimator. The loss is
\begin{equation}\label{lossf}
\mathcal L_{\text{noise}}(\theta)
=
\mathbb E_{\,t, x_0, \varepsilon}
\!\left[\, w(t)\,\big\lVert \varepsilon - E_\theta(x_t, t) \big\rVert_2^2 \right],
\end{equation}
where $w(t)>0$ is a per-step weight, and the diffusion step $t$ is sampled from a discrete distribution $t \sim p(t)$ on $\{1,\dots,T\}$. 

\subsection{Decoder Stage Activations}\label{act}
Let $E_\theta$ denote the trained U-Net denoiser. During inference for representation extraction, we freeze $E_\theta$ so it acts purely as a \textit{feature extractor}. Then, for a chosen noise level $t$, we construct $x_t$ from $x_0$ via the forward noising process and write $x_t(x_0)$ to make the dependence on the original image explicit. We feed both $x_t(x_0)$ and the same $t$ to $E_\theta$, ensuring \textit{the conditioning matches the corruption level whose noise the model was trained to predict}. Throughout, we treat $t$ as an explicit conditioning variable: features are always extracted from the denoiser under the same noise level at which the input was corrupted.

\paragraph{Readout Locations and Indexing.}
We extract decoder features after each residual block. The decoder contains four stages with different resolutions $\{16^2,32^2,64^2,128^2\}$ with three residual blocks per stage (RB1--RB3), giving 12 readout locations. We index these locations by a single flattened index $\ell \in \{1,\dots,12\}$, mapping $(r,b)$ to $\ell$ via $\ell = 3(r-1)+b$, where $r\in\{1,2,3,4\}$ denotes the decoder stage (from $16^2$ to $128^2$) and $b\in\{1,2,3\}$ denotes the residual block (RB1--RB3).

A decoder forward pass with inputs $(x_t(x_0), t)$ yields the readout activation at location $\ell$:
\begin{equation}
z^{(\ell)}(x_0, t) = u^{(\ell)}\!\big(x_t(x_0), t\big) \in \mathbb{R}^{C_\ell \times H_\ell \times W_\ell},
\end{equation}
where $u^{(\ell)}(\cdot,\cdot)$ denotes the corresponding activation mapping in the frozen \emph{decoder} of $E_\theta$, evaluated at $(x_t(x_0), t)$. Here $C_\ell, H_\ell, W_\ell$ are the channel count and spatial resolution of the feature tensor at readout location $\ell$. In particular, $H_\ell=W_\ell\in\{16,32,64,128\}$ depending on the decoder stage, and $C_\ell$ follows the decoder channel width at that stage.

To obtain a fixed-dimensional descriptor suitable for linear probing, we convert each feature tensor to a vector by spatial pooling. The corresponding embedding is obtained by \textit{global average pooling}:
\begin{equation}\label{finalrep}
\phi_{t,\ell}(x_0) = \operatorname{GAP}\!\big(z^{(\ell)}(x_0, t)\big) \in \mathbb{R}^{C_\ell}.
\end{equation}
This produces one feature vector per image, timestep, and readout location, enabling a controlled comparison of discriminative quality across the $(t,\ell)$ grid while keeping the downstream classifier capacity fixed.

\paragraph{Attention Placement.}
We insert a single self-attention module at the $16^2$ decoder stage (also in the encoder). At that stage, the final residual block includes this attention module, and we read features after it. All other readout locations correspond to purely convolutional residual blocks. Concretely, this design ensures that one readout location captures attention-augmented global interactions at an intermediate spatial grid, while the remaining readouts measure the contribution of convolutional processing alone at multiple resolutions.

\begin{figure*}
    \centering
    \includegraphics[width=0.85\linewidth]{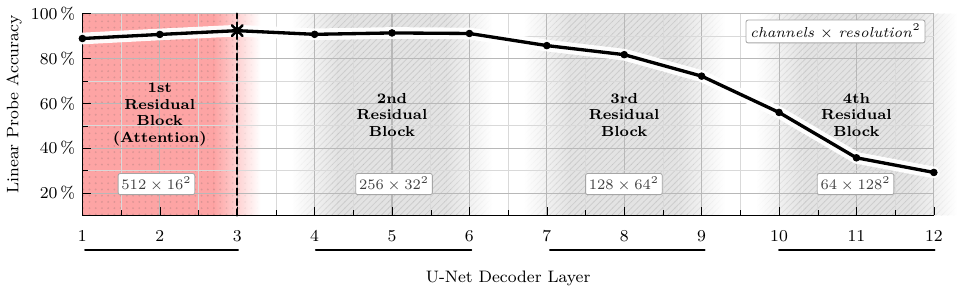}\vspace{-1mm}
    \caption{Linear probe accuracy across decoder readout locations $\ell \in \{1,\dots,12\}$ at the optimal noise level $t^\star = 25$. The DDPM runs for $T = 1000$ total timesteps, and we probe the subset $\{1, 10, 25, 50, 75, 100, 200, 400, 600\}$.}
    \label{fig:layer_sweep}\vspace{-2mm}
\end{figure*}

\paragraph{Why the Decoder?}
We place the attention readout in the decoder rather than the encoder or bottleneck for three reasons. First, decoder activations merge U-Net encoder features from the skip connections with conditioning on the current noise level $t$, producing representations that jointly capture local morphology and the progressively denoised global context most relevant for recognition. Second, at moderate and high $t$ the U-Net encoder is dominated by corrupted inputs, which lowers the signal-to-noise ratio for fine structures; in contrast, the decoder operates on features produced by the denoiser conditioned on noise level $t$, which are trained to predict denoising directions. Third, the bottleneck is deliberately narrow and global: it is well suited for coordinating the generative process but tends to suppress the part-level cues required for fine-grained discrimination. Empirically, this inductive bias is consistent with the linear probe results in Section~\ref{optimtime}, where decoder attention readouts yield the strongest separability across species.

\vspace{0.3mm}
\subsection{Linear Probe and Final Representation}
\label{sec:probe-cluster}
Using the decoder features $\phi_{t,\ell}$ defined in Equation \eqref{finalrep} as embeddings, we assess linear separability across $(t,\ell)$ via a \emph{linear probe} (linear softmax classifier) fitted independently at each $(t,\ell)$. With training data $\mathcal{D}_{\mathrm{tr}}=\{(x_{0,i},y_i)\}_{i=1}^{n}$ and test data $\mathcal{D}_\star=\{(x_{0,i}^\star,y_i^\star)\}_{i=1}^{n_\star}$, we set the embedding $\phi_{t,\ell}(x_0)$ and fit
$f_{t,\ell}(x_0) = W\,\phi_{t,\ell}(x_0) + b$ by minimizing the cross-entropy loss function
\begin{equation}
\mathcal{L}_{\mathrm{CE}}(W,b) = \frac{1}{n}\sum_{i=1}^{n} -\log\!\big(\mathrm{softmax}_{y_i}(f_{t,\ell}(x_{0,i}))\big).
\end{equation}
We evaluate $\operatorname{Acc}_\star(f_{t,\ell})$ on $\mathcal{D}_\star$ and choose $(t^\star,\ell^\star) \in \arg\max_{t,\ell}\operatorname{Acc}_\star(f_{t,\ell})$, i.e., the $(t,\ell)$ whose linear probe achieves the highest discriminative accuracy across layers and timesteps in the validation set. The final feature used for downstream tasks is then
\begin{equation}\label{finalrep1}
\phi(x_0) = \phi_{t^\star,\ell^\star}(x_0) \in \mathbb{R}^{C_{\ell^\star}}.
\end{equation}
The intuition behind these representations is to capture the most informative patterns that $E_\theta$ has learned when denoising at $t^\star$.

\vspace{0.3mm}
\subsection{Training Objective and Loss Weighting}
We train $E_\theta$ as an \emph{unconditional} denoiser on the training split using the objective in Equation ~\eqref{lossf}. The forward process is discretized into $T=1000$ diffusion steps with a cosine noise schedule. For the timestep weighting $w(t)$ we consider two variants: \vspace{-1mm}
\begin{align}
    w_{\text{MSE}}(t) &= 1, \\
    w_{\text{MinSNR}}(t) &= \frac{\min\bigl(\mathrm{SNR}(t), \gamma\bigr)}{\mathrm{SNR}(t)}.
\end{align}

Here $\operatorname{SNR}(t)$ denotes the \textit{signal-to-noise} ratio at diffusion step $t$ under the forward process, defined as
\begin{equation}
  \operatorname{SNR}(t) = \frac{\bar{\alpha}_t}{1 - \bar{\alpha}_t}.
\end{equation}
Unless otherwise stated, we train $E_\theta$ with MinSNR-$\gamma$ weighting $w_{\text{MinSNR}}(t)$, which downweights high-SNR (near-clean) timesteps relative to uniform MSE $w_{\mathrm{MSE}}(t)\equiv 1$ by setting $w_{\text{MinSNR}}(t)=\min(\mathrm{SNR}(t),\gamma)/\mathrm{SNR}(t)$.
\vspace{1mm}

Optimization uses AdamW (learning rate $5\cdot10^{-4}$, $\beta_1=0.9$, $\beta_2=0.999$, weight decay $10^{-4}$), cosine learning-rate decay with a $5\%$ warmup, global gradient clipping at $1.0$, mixed-precision training, and an exponential moving average of parameters with decay $0.999$. We train for $250$ epochs with a batch size of $256$ and select the checkpoint with the lowest validation loss (epoch 100). Class labels are \emph{never} used during denoiser training.

\section{Experiments \& Results}\label{results}
We evaluate our method on two datasets formed by Imaging FlowCytobot (IFCB) images from the SMHI \cite{torstensson2024smhiifcb} and SYKE \cite{kraft2022towards} Baltic programs. After preprocessing (see Appendix~\ref{app:data}), we obtain (i) a balanced dataset with $k=70$ classes, where each class contributes $500$ single-channel grayscale images resized to $128\times 128$ pixels, and (ii) an unbalanced dataset with $120$ classes and varying numbers of images per class. For each dataset, we construct class-stratified training and validation splits and keep a separate held-out split. The diffusion model is trained only on the training split. The validation split is used to track the diffusion loss, monitor generative quality via Fr\'echet Inception Distance (FID) \cite{heusel2017gans}, computed by replicating grayscale ROIs to three channels and using the standard Inception-V3 feature extractor, and select the network layer and diffusion timestep for representation extraction by linear probing; \emph{labels are never used during diffusion training}. The held-out test split, a fixed non-stratified subset without data augmentation, is used only for final downstream evaluation, where we report test accuracy and Macro F1 from the linear probe using the layer–timestep pair chosen on the validation set. Complete implementation details, including the diffusion-model architecture, linear probe configuration, and all hyperparameters, are provided in Appendix~\ref{app:impl}.

\subsection{Linear Probe Discriminant Analysis} \label{optimtime}
\vspace{-0.25mm}
To select the final diffusion representation, we sweep the U-Net decoder readout locations $\ell$ and denoising timesteps $t\in\{1,10,25,50,75,100,200,400,600\}$. For each pair $(t,\ell)$ we extract activations as defined in Section~\ref{sec:probe-cluster}, fit a linear softmax classifier on the corresponding training representations, and use validation accuracy to select $(t^\star,\ell^\star)$ following Section~\ref{sec:probe-cluster}. The best-performing probe is obtained for $t^\star=25$ at the decoder readout location $\ell^\star=3$ (RB3 at resolution $16^2$ under our flattening), so $(t^\star,\ell^\star)=\mathbf{(25,3)}$. This indicates that features at this stage already support near-linear class separation, combining global context from self-attention with a compact spatial grid ($16\times16$) and high channel capacity (512). As shown in Figure~\ref{fig:layer_sweep}, probe accuracy decreases for later decoder blocks (higher spatial resolution, lower channel count), suggesting subsequent upsampling prioritizes synthesis over linear separability. Full linear probe results are in Appendix~\ref{app:ext}.

\vspace{-0.25mm}
\subsection{Baselines}
\vspace{-0.25mm}
To contextualize our approach, we compare against supervised and self-supervised baselines. Supervised models include ResNet-50 \cite{he2016deep}, EfficientNet-B0/B3/V2-S \cite{tan2019efficientnet}, and ViT-B/16 \cite{dosovitskiy2020image} trained end-to-end with cross-entropy on the target training split, as well as a lightweight MLP on hand-crafted features (SIFT \cite{lowe2004sift} and edges). For self-supervised baselines, we start from ImageNet-1k pretrained MAE \cite{he2022masked} and DINOv3 \cite{simeoni2025dinov3} (ViT-B/16) and fine-tune the backbones on the target training images using their self-supervised objectives (no labels). We then freeze these fine-tuned backbones and train a linear probe on the target training split; the validation split is used for model selection (including any probe hyperparameters), and we report final results on the held-out test split. Our diffusion model is trained unconditionally from scratch on the target training images without labels and evaluated with the same frozen-feature linear-probe protocol. In Table~\ref{tab:supervised_vs_unsupervised_by_dataset}, we report the resulting metrics and parameter counts.

\subsection{Distribution Shift}

\setlength{\tabcolsep}{8pt}
\renewcommand{\arraystretch}{0.95}
\begin{table*}
\caption{Linear probe accuracy and Macro F1 on our balanced/unbalanced datasets for supervised baselines, self-supervised models, and our diffusion-based representation. Bold indicates the best result within groups, and $\uparrow$ ($\downarrow$) indicate metrics where higher (lower) is better.}

  \label{tab:supervised_vs_unsupervised_by_dataset}
  \centering
  \begin{tabular}{@{}lccccc@{}}
    \toprule
    & \multicolumn{2}{c}{\textbf{Balanced Dataset}} & \multicolumn{2}{c}{\textbf{Unbalanced Dataset}} & \\
    \cmidrule(lr){2-3}\cmidrule(lr){4-5}
    \textbf{Model} & \textbf{Accuracy} $\uparrow$ & \textbf{Macro F1} $\uparrow$ & \textbf{Accuracy} $\uparrow$& \textbf{Macro F1} $\uparrow$ & \textbf{Params} $\downarrow$\\
    \midrule
    \multicolumn{6}{l}{\textit{Supervised Methods}} \\
    \hspace{5mm} MLP on features     & 0.7602 & 0.7206 & 0.8651 & 0.7027 & \textbf{0.1M} \\
    \hspace{5mm} ResNet-50           & 0.9281 & 0.9229 & 0.9759 & 0.8864 & 25.5M \\
    \hspace{5mm} EfficientNet-B0     & 0.9490 & \textbf{0.9455} & \textbf{0.9761} & 0.8994 & 4M \\
    \hspace{5mm} EfficientNet-B3     & \textbf{0.9508} & 0.9447 & 0.9613 & 0.8421 & 11M \\
    \hspace{5mm} EfficientNet-V2-S   & 0.9302 & 0.9257 & 0.9100 & \textbf{0.9050} & 22.5M \\
    \hspace{5mm} ViT-B/16            & 0.9209 & 0.9128 & 0.9050 & 0.8980 & 87M \\
    \midrule
    \multicolumn{6}{l}{\textit{Self-Supervised Methods}} \\
    \hspace{5mm} Masked Autoencoders (ViT-B/16) & 0.8467 & 0.8214 & 0.8207 & 0.8001 & 87M \\
    \hspace{5mm} DINOv3 (ViT-B/16)   & 0.8031 & 0.6145 & 0.7909 & 0.3274 & 87M \\
    \hspace{5mm} \textit{\textbf{Our Diffusion}}       & \textbf{0.9240} & \textbf{0.9114} & \textbf{0.9145} & \textbf{0.9064} & \textbf{57M} \\
    \bottomrule
  \end{tabular}\vspace{-1mm}
\end{table*}

We assess out-of-distribution generalization using two unseen IFCB datasets. The first dataset \cite{kraft2022towards} is a year-long Baltic Sea collection acquired over a different 12-month period than our training data, spanning a broad range of seasonal and environmental conditions. The second is WHOI-22 \cite{Orenstein2015WHOIPlankton} from the U.S.\ Atlantic coast, which differs geographically and taxonomically. In all cases, we freeze the diffusion backbone and train only a linear classifier on top of the extracted features using the target dataset’s training split, and we report performance on the corresponding target test split. No target-domain images or labels are used to train the diffusion backbone; only the linear classifier is trained using target labels.

Our features exhibit strong transfer under both types of distribution shift, as shown in Table~\ref{tab:distribution-shift}. Overall accuracy is largely driven by the most frequent taxa, whereas Macro F1 averages performance across classes and is therefore much more sensitive to rare taxa. Because both Baltic (OOD) and WHOI-22 are naturally long-tailed, with several rare or weakly represented classes, accuracy can remain relatively high even when Macro F1 is lower, since reduced recall and precision on minority taxa has a limited effect on accuracy. This behavior is visible in Table~\ref{tab:distribution-shift}: performance remains robust under both temporal shift (Baltic) and geographic shift (WHOI-22), with a larger decrease in Macro F1 under the more challenging WHOI-22 shift.

\setlength{\tabcolsep}{3.5pt}
\renewcommand{\arraystretch}{0.95}
\begin{table}[H]
\caption{Out-of-distribution linear-probe evaluation on frozen diffusion features. The training regime denotes which source dataset was used to train the diffusion model weights. Metrics are computed on each target dataset’s label set. }

\centering
\begin{tabular}{lccc}
\toprule
\textbf{Dataset} & \textbf{Training regime} & \textbf{Accuracy} & \textbf{Macro F1} \\
\midrule
\multirow{2}{*}{Baltic (OOD)} & Balanced   & 0.9133 & 0.7356 \\
                              & Unbalanced & 0.9112 & 0.7564 \\
\midrule
\multirow{2}{*}{WHOI-22}      & Balanced   & 0.8411 & 0.8392 \\
                              & Unbalanced & 0.8272 & 0.8260 \\
\bottomrule
\end{tabular}

\label{tab:distribution-shift}
\end{table}

For OOD evaluation, we freeze the diffusion backbone trained on the source regime and extract features for all target images. We then train a linear classifier on the target dataset’s training split using the target labels and evaluate on the target test split. We do not restrict evaluation to only classes shared with the source training set, \textit{since the backbone can produce features for any input image} and the linear probe can be trained to predict the target labels from these frozen features. This protocol isolates representation transfer under distribution shift because the diffusion model weights remain fixed and only the lightweight linear probe is fit on the target domain. Moreover, because target class frequencies can differ substantially from the source regimes, this evaluation reflects performance under realistic long-tailed distributions.

\begin{figure*}
    \centering
    \includegraphics[width=0.965\linewidth]{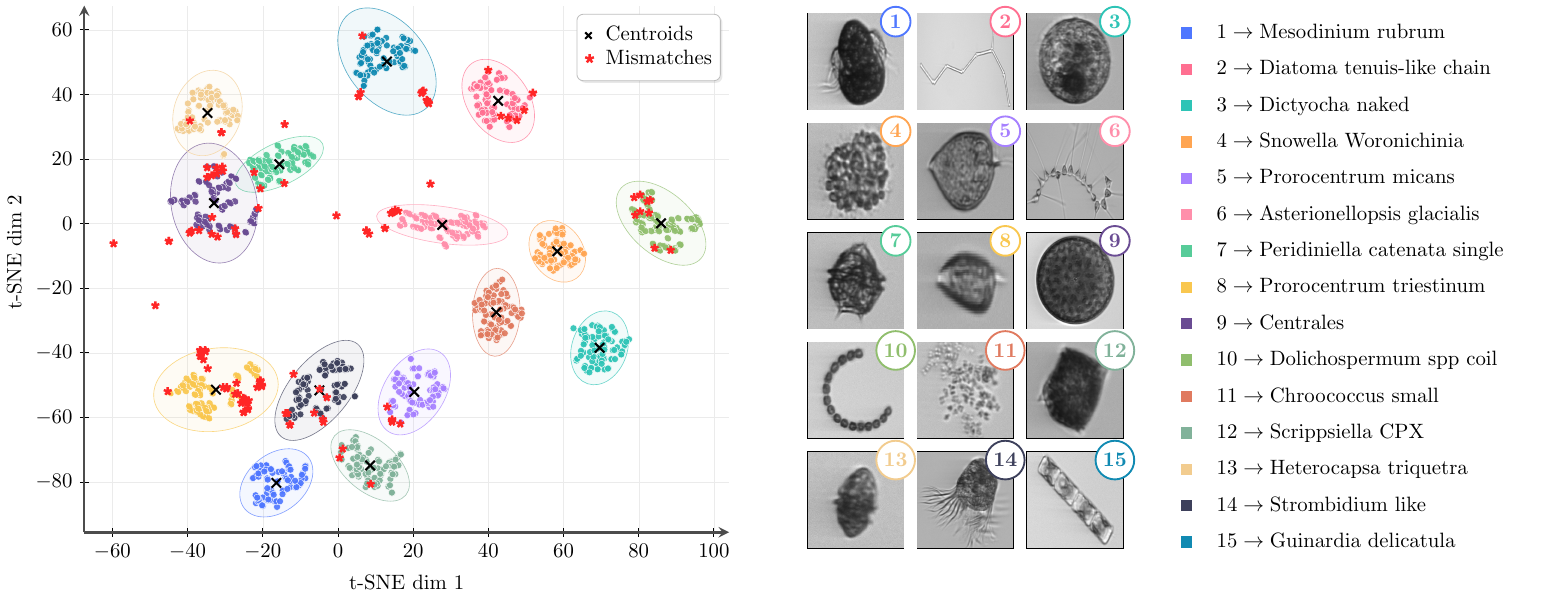}
    \caption{Visualization of learned representations via t-SNE with $k$-means (15 out of the 70 classes), with clusters matched to labels using the \textit{Hungarian algorithm} \cite{kuhn1955hungarian}. Mismatches represent points of the selected labels that did not fall under the correct cluster.}
    \label{tsne}\vspace{-3mm}
\end{figure*}

\subsection{Ablation Studies}
\begin{figure}[H]
    \centering
    \includegraphics[width=0.95\linewidth]{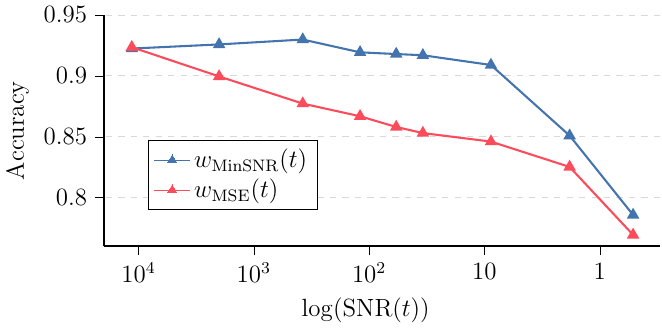}
    \caption{\textbf{Effect of loss on linear probe accuracy across noise.} Linear probe accuracy as a function of $\log \operatorname{SNR}(t)$ for the best decoder readout $(t^\star, \ell^\star)$. We compare two denoisers trained with identical architecture and schedule but different timestep weightings: uniform MSE weighting $w_{\text{MSE}}(t) \equiv 1$ and MinSNR ($\gamma$=5) weighting $w_{\text{MinSNR}}(t)$. High SNR corresponds to nearly clean inputs, low SNR to heavily noised inputs. Balanced dataset.\vspace{-2.25mm}
    }
    \label{fig:loss}
\end{figure}

\paragraph{Loss Function Ablation.}
We study how the training objective shapes the linear probe quality of diffusion features across noise levels. Specifically, we consider loss function \eqref{lossf} objectives that differ only in how they weight timesteps $t$ in the denoising loss. The first is the standard uniform MSE weighting, $w_{\text{MSE}}(t) \equiv 1$, and the second is the weighting $w_{\text{MinSNR}}(t)$  \cite{hang2023efficient}. Under uniform MSE, the mean-squared error allocates substantial weight to both very easy high-SNR and hard low-SNR conditions, which can under-train the mid-SNR regime where much of the denoising trajectory resides. In contrast, MinSNR-$\gamma$ weighting with $\gamma = 5$ counteracts this by capping the contribution of extreme SNRs and emphasizing the informative middle of the schedule. As shown in Figure~\ref{fig:loss}, features learned with MinSNR-$\gamma$ maintain classification accuracy as SNR decreases, and in the high-SNR (early) regime they show a slight improvement, whereas on our IFCB plankton experiments, features learned with uniform MSE degrade steadily.

\vspace{-3mm}
\paragraph{Generation Fidelity.} During training we monitored both FID and validation loss and found that they decouple: after a certain point, validation loss shows clear overfitting while FID continues to improve.  This behavior arises because FID measures proximity between generated samples and the validation distribution, which in our setting is close to the training distribution; as the generator memorizes training statistics, sample quality by FID can keep rising even as the model overfits and its representations degrade. To verify the effect on downstream utility, we compared an overfit checkpoint with a better FID to a checkpoint selected by the minimum validation loss. Linear probing on frozen features revealed an accuracy drop from 0.9240 to 0.8812, roughly a four-point absolute decline, for the overfit model. These results, detailed in Appendix~\ref{app:ext}, suggest that FID is useful as a diagnostic but should not be the optimization target when training diffusion models for representation learning on fine-grained classification tasks.

\subsection{Clustering as a Diagnostic} 
\vspace{-0.25mm}
Beyond the linear probe, we examine class-consistent structure by running $k$-means (balanced dataset) with $k=70$ on \emph{test} embeddings $\phi(x_{0})$ extracted at $(t^\star,\ell^\star)$. This analysis complements the linear probe, which directly evaluates linear separability, \textit{by probing structure in the absence of supervision} \cite{alfano2022efficient}. Figure~\ref{tsne} presents a 2D t-SNE embedding of the learned representations for a subset of visually similar plankton classes, annotated with $k$-means partitions, to assess fine-grained class separability. The visualization reveals both well-separated and overlapping regions, highlighting where the model confuses classes and how the feature space organizes taxonomically coherent groupings. A more comprehensive set of clustering metrics is provided in  Appendix~\ref{app:ext}.

\vspace{-1mm}
\section{Discussions}\label{conclu}
\vspace{-0.25mm}
We extract representations from a frozen diffusion backbone by reading decoder activations at multiple depths and noise timesteps. For each $(t,\ell)$, we train a linear classifier to measure linear separability and select the pair that maximizes validation accuracy. Across our settings, the most discriminative features consistently arise in early decoder readouts at small-to-moderate noise levels, where the representation retains intermediate-level shape and texture while suppressing overly fine, instance-specific details. This aligns with the intuition that high noise preserves only coarse structure, whereas very low noise reintroduces fine-scale variation that is less stable for linear separation. In our experiments, the best probe accuracy is obtained at the selected readout corresponding to the final attention-equipped decoder block, suggesting that late attention refines class-relevant cues without collapsing the representation into sample-specific detail.

Across experiments, the MinSNR-$\gamma$ objective shifts learning toward slightly noisier timesteps, producing more noise-robust, morphology-preserving representations and improving linear probing on fine-grained plankton images. We also find that higher generative quality doesn’t necessarily translate to better discriminative performance, so the two should be evaluated separately.

These observations yield a practical guideline for our setting: use a single early to mid layer feature at low noise from a frozen backbone and train only a linear probe. This keeps adaptation lightweight, reduces instrument-specific overfitting, and supports transfer by reusing the same features across datasets. Although diffusion training is costlier than CNNs, the cost is amortized by reusing the backbone and only refitting the probe per dataset.

Our experiments use IFCB plankton images, but the feature extractor is self-supervised and makes no instrument- or dataset-specific assumptions. It learns from raw images without microscopy rules, hand-crafted features, or acquisition priors; labels are used only by the downstream classifier. As a result, the same representation supports competitive linear probes compared with strong supervised baselines and clearly outperforms ViT-based self-supervised baselines, and it transfers to other labeled plankton collections and environmental microscopy datasets.

\vspace{-3mm}
\paragraph{Are Diffusion Models Learning Morphology?}
When extracting representations, we use the decoder block selected by the linear probe and treat all its spatial locations across a batch of plankton images as feature tokens. Running PCA on these tokens and mapping the first three components to RGB yields the overlays in Figure~\ref{fig:PCA_visualization}, where consistent colors along each organism across noise levels suggest that diffusion features are organized by plankton morphology rather than raw pixel values.

\paragraph{Runtime and Memory.}
We train our diffusion model for 15~h on \(4\times\) NVIDIA A100-40GB GPUs on a single node. Peak GPU memory usage is \(\sim 38\)~GB per GPU. Training uses 24 CPU threads and \(\approx 11\)~GB of disk space for checkpoints and logs.

\vspace{-3mm}
\paragraph{Limitations \& Extensions.}\label{future_work}
A limitation of this study is that the above findings on layer--timestep selection and the location of the most discriminative decoder features are established for a single U-Net family, a fixed attention placement, and the noise schedules we evaluate on IFCB plankton imagery; extending the same sweep to additional architectures, schedules, and datasets would be required to claim broader stability. This work is also limited to a single Baltic site and grayscale ROI imagery, so conclusions are environment- and acquisition-specific.

Data-wise, next steps are to test whether diffusion-based representations outperform standard CNNs on heavily imbalanced data and to quantify how augmentation reduces imbalance by stratifying noise-prediction loss by augmentation, probing timestep- and decoder-wise invariances with linear probes and $k$-means retrieval in a shared embedding space, and using padding diagnostics to separate biological signal from instrument artifacts.

Model-wise, directions include pretraining and fine-tuning, cross-instrument adaptation, and alternative diffusion designs such as latent diffusion and EDM-style schedules \cite{karras2022elucidating}; ablating the timestep-sampling distribution $p(t)$ (we use squared-cosine importance sampling) and its interaction with MinSNR-$\gamma$ to disentangle sampling effects from loss reweighting, for example by comparing against uniform timestep sampling; replacing FID with metrics that disentangle fidelity from coverage; tracking noise-prediction loss across timesteps and relating it to held-out quality to distinguish structure learning from texture memorization and detect overfitting; recognizing that clustering uses representations chosen with supervised timestep/layer selection and is therefore not a label-free baseline; and studying concatenated multi-step features, attention placement, and linear probe plus attribution analyses beyond raw attention to localize class-separable, task-relevant signals.

{
    \small
    \bibliographystyle{ieeenat_fullname}
    \bibliography{main}
}

\clearpage
\onecolumn
\appendix

\setlength{\parindent}{0pt}
\setlength{\parskip}{4.5pt plus 2pt minus 1pt} 

\section{Dataset Details}\label{app:data}
In this section, we describe how we construct the plankton datasets used in our experiments. We begin by integrating the raw, separately annotated datasets obtained from the \textbf{SYKE} \cite{kraft2022towards} and \textbf{SMHI} \cite{torstensson2024smhiifcb} Baltic monitoring programs, which sample plankton communities along the \textbf{Swedish coast}, into a single label space spanning all taxa. From this common label space, we then define the in-distribution datasets used for training and evaluation, including both a balanced and an unbalanced variant, and we present the two out-of-distribution datasets used to assess model generalization. We first detail the curation and class-harmonization procedure, explaining how taxonomic labels from the two sources are merged into common categories and how these categories are used to construct the balanced and unbalanced datasets. We conclude by describing the preprocessing pipeline applied to all images. All taxonomic assignments follow the World Register of Marine Species (WoRMS) \cite{worms_specimens_2025}, which we adopt as a standardized reference for marine organism nomenclature.

A central characteristic of these datasets is the variability in taxonomic resolution. The included class labels span \textbf{multiple taxonomic ranks}, ranging from \textit{species} to \textit{genus} and even \textit{order}. This variation stems from the differing levels of detail visible in plankton images: some organisms display clear morphological features that enable confident species-level identification, while others are ambiguous and can only be classified at broader levels. This is especially relevant when using automated imaging systems such as the Imaging FlowCytobot (IFCB) instruments used by SYKE and SMHI.

To accommodate this variability, the dataset includes both precise labels (e.g., defined species names) and generalized taxonomic categories using the standard \textbf{``sp.''} and \textbf{``spp.''} conventions. These indicate that the exact species is unknown: \textit{Genus sp.} refers to a single, unidentified species within a genus, while \textit{Genus spp.} refers to a group of species within the same genus. These broader categories ensure that valuable samples are not excluded due to taxonomic uncertainty.

\subsection{Source Data and Taxonomic Harmonization}\label{app:data1}
We construct our datasets from IFCB region-of-interest (ROI) image catalogs curated by the SMHI and SYKE Baltic monitoring programs. IFCB instruments acquire grayscale ROI images, i.e., cropped fields of view centered on individual particles (such as plankton cells, colonies, or detrital aggregates), along with expert-provided taxonomic labels maintained in each provider’s taxonomy.

To obtain consistent supervision across providers, we first harmonize the SMHI and SYKE taxonomies by collapsing synonymous labels and resolving provider-specific naming conventions. From the intersection of these harmonized taxonomies, we derive two datasets that share the same preprocessing pipeline but differ in class-frequency structure (balanced vs.\ naturally long-tailed). 

The merged dataset was constructed by integrating four annotated image datasets: three provided by SMHI and one by SYKE.

\begin{itemize}[itemsep=8pt]
    \item \begin{minipage}[t]{\linewidth}\RaggedRight\ttfamily smhi\_ifcb\_baltic\_annotated\_images (1)\end{minipage}
    \item \begin{minipage}[t]{\linewidth}\RaggedRight\ttfamily smhi\_ifcb\_skagerrak\_kattegat\_annotated\_images (2)\end{minipage}
    \item \begin{minipage}[t]{\linewidth}\RaggedRight\ttfamily smhi\_ifcb\_t\_ngesund\_annotated\_images (3)\end{minipage}
    \item \begin{minipage}[t]{\linewidth}\RaggedRight\ttfamily phytoplankton\_labeled\_SYKE (4)\end{minipage}
\end{itemize}

The three SMHI datasets (1-3) were first merged by aligning their folder structures and class names, which followed a consistent naming convention. After this unified SMHI dataset was created, the SYKE dataset (4) was merged in. While several class names matched exactly and were merged directly, others required taxonomic harmonization or renaming. The following decisions were taken during this process:

\begin{itemize}[itemsep=8pt]
    \item \begin{minipage}[t]{\linewidth}\RaggedRight\ttfamily Gonyaulax\_verior (4) was merged with Gonyaulax\_spp (2) under the name Gonyaulax\_verior.\end{minipage}
    \item \begin{minipage}[t]{\linewidth}\RaggedRight\ttfamily Licmophora\_sp (4) was merged with Licmophora (2) under Licmophora\_sp.\end{minipage}
    \item \begin{minipage}[t]{\linewidth}\RaggedRight\ttfamily Nodularia\_spumigena (4), which included both coiled and filamentous forms, was merged with Nodularia\_spumigena\_coil and Nodularia\_spumigena\_filament (1). The resulting class was named Nodularia\_spumigena\_coil.\end{minipage}
    \item \begin{minipage}[t]{\linewidth}\RaggedRight\ttfamily centrales\_sp (4) was merged with Centrales (1,2). Notably, SYKE samples sometimes contained multiple planktonic elements, while SMHI samples generally showed single elements.\end{minipage}
    \item \begin{minipage}[t]{\linewidth}\RaggedRight\ttfamily Oocystis\_sp (4) was merged with Oocystis\_spp (1) under Oocystis\_sp.\end{minipage}
    \item \begin{minipage}[t]{\linewidth}\RaggedRight\ttfamily Monoraphidium\_contortum (4) was merged with Monoraphidium\_spp (1) under Monoraphidium\_contortum.\end{minipage}
    \item \begin{minipage}[t]{\linewidth}\RaggedRight\ttfamily Pennales\_sp\_thick (4) was merged into Pennales.\end{minipage}
    \item \begin{minipage}[t]{\linewidth}\RaggedRight\ttfamily Peridiniella\_catenata (1) was merged with Peridiniella\_catenata\_chain and Peridiniella\_catenata\_single (4).\end{minipage}
    \item \begin{minipage}[t]{\linewidth}\RaggedRight\ttfamily Eutreptiella\_sp (4) was merged with Eutreptiella\_spp (1,2,3) under Eutreptiella\_spp.\end{minipage}
    \item \begin{minipage}[t]{\linewidth}\RaggedRight\ttfamily Cyclotella\_choctawhatcheeana (4) was merged into Cyclotella\_spp (1) under Cyclotella\_choctawhatcheeana.\end{minipage}
    \item \begin{minipage}[t]{\linewidth}\RaggedRight\ttfamily Chaetoceros\_sp (4) was merged with Chaetoceros\_spp (3).\end{minipage}
    \item \begin{minipage}[t]{\linewidth}\RaggedRight\ttfamily Chaetoceros\_sp\_single (4) was merged with Chaetoceros\_single\_cell (1).\end{minipage}
    \item \begin{minipage}[t]{\linewidth}\RaggedRight\ttfamily ciliata (4) was merged with ciliates (2), adopting the taxonomically correct label Ciliata according to WoRMS.\end{minipage}
\end{itemize}

\subsection{Balanced and Unbalanced Datasets}

From the curated dataset obtained in Section~\ref{app:data1}, we derive two datasets that we use to train our diffusion models.

\paragraph{Balanced dataset.}
To obtain a controlled benchmark focused on representation quality, we first select all classes that have at least $100$ labeled images in the curated dataset, which leaves us with $k = 70$ classes. For each of these $k$ classes, we then construct a dataset by randomly sampling up to $500$ single-channel grayscale images per class. We first perform a class-stratified split into training, validation, and test sets with an $80/10/10$ ratio, yielding up to $400$, $50$, and $50$ images per class, respectively. We then enforce class balance in the training split by ensuring that each class contributes exactly $400$ samples. For classes that had fewer than $500$ samples in the original curated dataset, we obtain the missing training samples by augmenting only the training split, leaving the validation and test sets unaugmented. Each augmented training image is generated by applying a single geometric transformation at a time to a real training image: horizontal flip, vertical flip, $180^\circ$ rotation, or uniform color intensity scaling by $-10\%$, $-5\%$, $+5\%$, or $+10\%$.

\vspace{-2mm}
\paragraph{Unbalanced (long-tailed) dataset.}
To mirror realistic monitoring conditions, we construct an unbalanced dataset starting from the curated dataset. We first remove classes with fewer than $5$ labeled images, as such classes are too small to be informative for training or evaluation. After this filtering step, we are left with $120$ classes. For each of these classes, we then include all available images that pass the preprocessing and quality-control steps described in Section~\ref{app:data:preproc}, without imposing any further per-class subsampling. As in the balanced setting, we perform a class-stratified split into training, validation, and test sets using an $80/10/10$ ratio. However, unlike in the balanced case where we explicitly enforce $400$ training samples per class, here rare classes may contribute only a handful of training images, and even when we apply the same set of simple geometric augmentations used in the balanced case (horizontal flip, vertical flip, $180^\circ$ rotation, and uniform color intensity scaling by $-10\%$, $-5\%$, $+5\%$, or $+10\%$) only to the training split, their effective training size remains far below $400$ samples. As a result, the training set is heavily long-tailed, with a few abundant classes and many rare species, closely matching the operational scenario in automated plankton monitoring.

\vspace{-2mm}

\paragraph{Splits usage.}
The diffusion model is trained \emph{only} on the training split, without labels. The validation split is non-augmented and used to track diffusion loss, monitor generative quality via FID, and select the layer-timestep pair for the linear probe. The held-out test split is a fixed, non-augmented subset used exclusively for downstream evaluation, where we report test accuracy and Macro F1 for the linear classifier trained on frozen diffusion features.

\subsection{Out-of-Distribution Test Sets}

We assess distribution robustness on two unseen IFCB datasets by freezing the diffusion backbone and training a linear probe on an 80/20 target train/test split, then reporting Accuracy and Macro F1 on the target test set.

\vspace{-3mm}
\paragraph{Baltic (OOD).}
The first out-of-distribution (OOD) dataset is a year-long Baltic Sea collection from 2021 \cite{kraft2022towards}, acquired over a 12-month period that is disjoint from the years used for the SMHI and SYKE data employed to construct our balanced and unbalanced source datasets. It reflects the realistic distribution of Baltic plankton across the full annual cycle, from ice-affected winter conditions to spring and summer blooms, and thus provides a natural, unsorted, and robust testbed for validating our models under temporal and environmental shifts. For this dataset, we freeze the diffusion backbone  trained on either the 70-class balanced or the 120-class unbalanced source regime, extract features, and train only a linear classifier on Baltic labels. The diffusion model is never exposed to this Baltic collection during training, so it is used purely as an out-of-distribution test set. When mapping Baltic labels to the source taxonomies, all Baltic classes are covered by the 120-class unbalanced source dataset, whereas the 70-class balanced source dataset lacks three Baltic taxa: \texttt{Aphanothece\_paralleliformis}, \texttt{Licmophora\_sp}, and \texttt{Melosira\_arctica}.

\vspace{-3mm}
\paragraph{WHOI-22.}
The second OOD dataset, WHOI-22 \cite{Orenstein2015WHOIPlankton}, is a curated subset of the WHOI-Plankton benchmark acquired on the U.S.\ Atlantic coast using IFCB instrumentation. It differs from the Baltic data in geography, community composition, and acquisition conditions, and includes taxa not present in our Baltic source datasets. As with the Baltic (OOD) set, we evaluate by freezing the diffusion backbone trained on the Baltic balanced or unbalanced regime and fitting a linear classifier on top of the extracted features. As in the Baltic OOD dataset, the diffusion backbone is never exposed to target-domain images. When mapping WHOI-22 labels to the Baltic source taxonomies, several WHOI-22 taxa are not covered by one or both source regimes: the 120-class unbalanced dataset lacks \texttt{DactFragCeratul}, \texttt{Euglena}, \texttt{Phaeocystis}, \texttt{Pleurosigma}, \texttt{detritus}, \texttt{nanoflagellate}, \texttt{other\_lt20}, and \texttt{pennate}, and the 70-class balanced dataset additionally lacks \texttt{Asterionellopsis}, \texttt{Cylindrotheca}, \texttt{Dinobryon}, \texttt{Ditylum}, and \texttt{Licmophora\_sp}. The full overlap pattern is summarized in Table~\ref{tab:whoi22-overlap}.

\vspace{-1mm}
\renewcommand{\arraystretch}{0.82}
\begin{table}[ht!]
    \centering
    \caption{Taxa in WHOI-22 that are \emph{not covered} by at least one of the two Baltic source taxonomies (120-class unbalanced and 70-class balanced). ``Yes'' indicates that the taxon is present in the corresponding source dataset; ``No'' indicates it is absent.}

    \begin{tabular}{lcc}
        \toprule
        WHOI-22 taxon & 120-class unbalanced & 70-class balanced \\
        \midrule
        \texttt{Asterionellopsis}   & Yes & No \\
        \texttt{Cylindrotheca}      & Yes & No \\
        \texttt{DactFragCeratul}    & No  & No \\
        \texttt{Dinobryon}          & Yes & No \\
        \texttt{Ditylum}            & Yes & No \\
        \texttt{Euglena}            & No  & No \\
        \texttt{Licmophora\_sp}     & Yes & No \\
        \texttt{Phaeocystis}        & No  & No \\
        \texttt{Pleurosigma}        & No  & No \\
        \texttt{detritus}           & No  & No \\
        \texttt{nanoflagellate}     & No  & No \\
        \texttt{other\_lt20}        & No  & No \\
        \texttt{pennate}            & No  & No \\
        \bottomrule
    \end{tabular}
    \label{tab:whoi22-overlap}
\end{table}

\vspace{-2.5mm}
\subsection{Preprocessing Pipeline}\label{app:data:preproc}
\vspace{-0.5mm}
All curation and preprocessing precede any model training and are applied identically to the training, validation, test, and OOD partitions. The pipeline mapping raw IFCB ROIs to standardized recognition images is fully deterministic.

\begin{enumerate}
  \item \textbf{File integrity and metadata checks.} We discard corrupt images and zero-sized files. Each class name is assigned to a unique integer class identifier to ensure one-to-one label indices across SMHI and SYKE.
  \item \textbf{Grayscale standardization.} IFCB images are inherently single-channel. Any multi-channel encodings are converted to a single channel via luminance, and intensities are linearly rescaled to $[0, 1]$.
  \item \textbf{Resizing.} Each ROI is resized to a $128 \times 128$ square without padding. In preliminary experiments, we evaluated several padding strategies, but pure resizing yielded the most reliable downstream performance for our U-Net-based diffusion model, so we adopt it throughout. However, a systematic study of padding and resizing schemes for plankton data remains an open challenge.
\end{enumerate}

\subsection{Augmentations}
Across both the balanced and unbalanced source regimes, we adopt a deliberately conservative augmentation strategy that is applied \emph{only} to the training split. In both cases, augmented samples are generated from real ROIs using the same small set of label-preserving geometric transformations: horizontal flip, vertical flip, $180^\circ$ rotation, and uniform color intensity scaling by $-10\%$, $-5\%$, $+5\%$, or $+10\%$, with a single transformation applied per original image. In the balanced regime, this allows us to synthetically increase underrepresented classes so that each class contributes exactly $400$ training examples, while leaving the validation and test splits unchanged. In the unbalanced, long-tailed regime, the same augmentations modestly strengthen supervision for rare taxa without coming close to equalizing class frequencies, so the training distribution remains heavily long-tailed.

We restrict ourselves to these simple geometric operations because they encode natural invariances of IFCB imagery: the orientation of the organism in the flow cell is arbitrary, and slight changes in apparent size are common due to focus and segmentation variability, whereas more aggressive manipulations (such as strong cropping or elastic deformations) could distort morphological cues that are important for taxonomic identity. Applying augmentations exclusively to the training split ensures that the validation, test, and OOD datasets consist solely of real, preprocessed ROIs, so that all reported performance reflects behavior on realistic acquisition conditions while still benefiting from an increased effective sample size during model training.

\renewcommand{\arraystretch}{0.95}
\begin{table}[h]
\centering
\small
\caption{List of the plankton species in our balanced dataset.}\vspace{1mm}
\setlength{\tabcolsep}{8pt}
\begin{tabular}{ll}
\toprule
Alexandrium\_pseudogonyaulax & Gymnodinium\_like \\
Aphanizomenon\_flosaquae & Heterocapsa\_rotundata \\
Aphanizomenon\_spp\_bundle & Heterocapsa\_triquetra \\
Aphanizomenon\_spp\_filament & Heterocyte \\
Asterionellopsis\_glacialis & Leptocylindrus\_danicus \\
Beads & Leptocylindrus\_danicus\_minimus \\
Centrales & Lingulodinium\_polyedrum \\
Cerataulina\_pelagica & Mesodinium\_rubrum \\
Chaetoceros\_chain & Monoraphidium\_contortum \\
Chaetoceros\_single\_cell & Nodularia\_spumigena \\
Chaetoceros\_spp & Octactis\_speculum \\
Chroococcales & Oocystis\_sp \\
Chroococcus\_small & Oscillatoriales \\
Ciliata & Pauliella\_taeniata \\
Cryptomonadales & Pennales\_sp\_thick \\
Cryptophyceae\_Teleaulax & Pennales\_sp\_thin \\
Cyclotella\_choctawhatcheeana & Peridiniella\_catenata\_chain \\
Cylindrotheca\_Nitzschia\_longissima & Peridiniella\_catenata\_single \\
Cymbomonas\_tetramitiformis & Prorocentrum\_cordatum \\
Dactyliosolen\_fragilissimus & Prorocentrum\_micans \\
Diatoma\_tenuis-like\_chain & Prorocentrum\_triestinum \\
Diatoma\_tenuis-like\_single\_cell & Pseudo\_nitzschia\_spp \\
Dictyocha\_naked & Pseudopedinella\_sp \\
Dino\_smaller\_than\_30unidentified & Pseudosolenia\_calcar\_avis \\
Dinophyceae & Pyramimonas\_sp \\
Dinophysis\_acuminata & Rhizosolenia\_Pseudosolenia \\
Dolichospermum\_Anabaenopsis & Scrippsiella\_CPX \\
Dolichospermum\_Anabaenopsis\_coiled & Scrippsiella\_group \\
Dolichospermum\_spp\_coil & Skeletonema\_marinoi \\
Dolichospermum\_spp\_filament & Snowella\_Woronichinia \\
Euglenophyceae & Strombidium\_like \\
Eutreptiella\_spp & Thalassiosira\_gravida \\
Guinardia\_delicatula & Thalassiosira\_levanderi \\
Gymnodiniales & Tripos\_lineatus \\
Gymnodiniales\_smaller\_than\_30 & Uroglenopsis\_sp \\
\bottomrule
\end{tabular}
\label{tab:plankton_species_clean2}
\end{table}

\renewcommand{\arraystretch}{0.95}
\begin{table}[H]
\centering
\small
\caption{List of the plankton species in our unbalanced dataset, including the number of samples per class.}\vspace{1mm}
\setlength{\tabcolsep}{8pt}
\begin{tabular}{llll}
\toprule
Alexandrium\_pseudogonyaulax & 387 & Gyrodinium\_spirale & 65 \\
Amphidinium\_like & 40 & Gyrosigma\_Pleurosigma & 40 \\
Amylax\_triacantha & 146 & Heterocapsa\_rotundata & 550 \\
Apedinella\_radians & 307 & Heterocapsa\_triquetra & 550 \\
Aphanizomenon\_flosaquae & 550 & Heterocyte & 513 \\
Aphanizomenon\_spp\_bundle & 407 & Karenia\_mikimotoi & 315 \\
Aphanizomenon\_spp\_filament & 550 & Katablepharis\_remigera & 304 \\
Aphanothece\_paralleliformis & 211 & Katodinium\_like & 268 \\
Asterionellopsis\_glacialis & 550 & Leptocylindrus\_danicus & 380 \\
Beads & 375 & Leptocylindrus\_danicus\_minimus & 396 \\
Binuclearia\_lauterbornii & 297 & Licmophora\_sp & 325 \\
Centrales & 550 & Lingulodinium\_polyedrum & 550 \\
Cerataulina\_pelagica & 351 & Melosira\_arctica & 293 \\
Ceratoneis\_closterium & 295 & Merismopedia\_sp & 348 \\
Chaetoceros\_cf\_convolutus & 344 & Mesodinium\_major & 319 \\
Chaetoceros\_chain & 360 & Mesodinium\_rubrum & 550 \\
Chaetoceros\_danicus & 302 & Monoraphidium\_contortum & 535 \\
Chaetoceros\_single\_cell & 522 & Nitzschia\_paleacea & 315 \\
Chaetoceros\_spp & 550 & Nodularia\_spumigena & 537 \\
Chaetoceros\_wighamii & 195 & Octactis\_speculum & 363 \\
Chlorococcales & 345 & Oocystis\_sp & 550 \\
Chroococcales & 392 & Oscillatoriales & 550 \\
Chroococcus\_small & 550 & Pauliella\_taeniata & 369 \\
Ciliata & 534 & Pennales\_sp\_thick & 494 \\
Ciliophora & 315 & Pennales\_sp\_thin & 550 \\
Cryptomonadales & 550 & Peridiniales\_smaller\_than\_30 & 89 \\
Cryptophyceae\_Teleaulax & 550 & Peridiniella\_catenata\_chain & 478 \\
Cyclotella\_choctawhatcheeana & 462 & Peridiniella\_catenata\_single & 550 \\
Cylindrotheca\_Nitzschia\_longissima & 453 & Proboscia\_alata & 41 \\
Cymbomonas\_tetramitiformis & 449 & Prorocentrum\_cordatum & 532 \\
Dactyliosolen\_fragilissimus & 533 & Prorocentrum\_micans & 550 \\
Diatoma\_tenuis-like\_chain & 550 & Prorocentrum\_triestinum & 550 \\
Diatoma\_tenuis-like\_single\_cell & 545 & Pseudo\_nitzschia\_spp & 550 \\
Dictyocha\_fibula & 81 & Pseudopedinella\_sp & 538 \\
Dictyocha\_naked & 550 & Pseudosolenia\_calcar\_avis & 403 \\
Dino\_larger\_than\_30unidentified & 327 & Pyramimonas\_sp & 550 \\
Dino\_smaller\_than\_30unidentified & 544 & Rhizosolenia\_Pseudosolenia & 387 \\
Dinobryon\_spp & 332 & Rhizosolenia\_hebetata\_f\_semispina & 40 \\
Dinoflagellate\_smaller\_than\_30 & 146 & Rhizosolenia\_setigera & 298 \\
Dinophyceae & 550 & Scenedesmus\_spp & 162 \\
Dinophysis\_acuminata & 538 & Scrippsiella\_CPX & 550 \\
Dinophysis\_norvegica & 260 & Scrippsiella\_group & 517 \\
Ditylum\_brightwellii & 327 & Skeletonema\_marinoi & 550 \\
Dolichospermum\_Anabaenopsis & 550 & Snowella\_Woronichinia & 550 \\
Dolichospermum\_Anabaenopsis\_coiled & 550 & Strombidium\_like & 550 \\
Dolichospermum\_spp\_coil & 550 & Thalassionema\_nitzschioides & 203 \\
Dolichospermum\_spp\_filament & 550 & Thalassiosira\_anguste-lineata & 41 \\
Double\_cells & 307 & Thalassiosira\_gravida & 363 \\
Ebria\_tripartita & 41 & Thalassiosira\_levanderi & 550 \\
Enisiculifera\_carinata & 97 & Thalassiosira\_nordenskioeldii & 334 \\
Euglenophyceae & 352 & Thalassiosira\_punctigera & 73 \\
Eutreptiella\_spp & 550 & Thalassiosira\_spp & 308 \\
Gonyaulax\_spinifera & 81 & Torodinium\_robustum & 347 \\
Gonyaulax\_verior & 170 & Tripos\_furca & 146 \\
Guinardia\_delicatula & 550 & Tripos\_fusus & 57 \\
Guinardia\_flaccida & 57 & Tripos\_lineatus & 429 \\
Gymnodiniales & 532 & Tripos\_muelleri & 122 \\
Gymnodiniales\_smaller\_than\_30 & 474 & Uroglenopsis\_sp & 550 \\
Gymnodinium\_like & 408 & Warnowia\_like & 113 \\

\bottomrule
\end{tabular}
\label{tab:plankton_species_clean}
\end{table}

\section{Implementation and Reproducibility}\label{app:impl}
\vspace{-1mm}
To enable faithful replication and extension of our results, this section specifies the exact model components and training/evaluation settings we use. We describe implementation choices at the level needed to re-run experiments end-to-end, including architectural details and hyperparameters.

\vspace{-2mm}
\subsection{Full U-Net Architecture}\label{architecture}
\vspace{-1.5mm}
We use a DDPM-style U-Net denoiser $E_\theta$ implemented in \texttt{diffusers} (\texttt{UNet2DModel}) operating on single-channel $128{\times}128$ inputs. The U-Net encoder channel pattern is $[64,\,128,\,256,\,512]$; the decoder mirrors these widths with skip connections. Residual blocks are pre-activation with \texttt{GroupNorm(16)} and \texttt{SiLU}, and no cross-attention is used. We use \textit{self-attention} only at the $16\times 16$ stage (after the last residual block in both encoder and decoder), as this mid-resolution offers a good trade-off between receptive field, spatial detail, and downstream feature quality, in line with prior analyses of diffusion backbones and intermediate noise scales~\cite{fuest2024diffusion, dhariwal2021diffusion}.

\renewcommand{\arraystretch}{1.05}
\begin{table}[h]
\centering
\caption{U-Net configuration (Diffusers).}
\label{tab:unet-config}
\begin{tabular}{lp{9.5cm}}
\toprule
\textbf{Configuration} & \textbf{Value} \\
\midrule
Architecture & DDPM-style U-Net denoiser (\texttt{UNet2DModel}) \cite{vonplaten2022diffusers}. \\
Channel multipliers & $1{-}2{-}4{-}8$ (stages at $128^2,64^2,32^2,16^2$). \\
Attention resolutions & $\{16\}$ (self-attention only at $16{\times}16$--both encoder and decoder). \\
Blocks per resolution & encoder $=2$, bottleneck $=2$, decoder $=3$. \\
Normalization & GroupNorm(16). \\
Activation & SiLU. \\
Params & $\sim 57$M. \\
\bottomrule
\end{tabular}
\end{table}

\subsection{Training Details and Compute}
We optimize the $\varepsilon$-prediction objective presented in the main article with MinSNR-$\gamma$ weighting:
\begin{equation}
w(t)=\frac{\min(\mathrm{SNR}(t),\gamma)}{\mathrm{SNR}(t)},\ \ \gamma=5,
\end{equation}
which rebalances over/under-weighted timesteps and improves sample efficiency \cite{hang2023efficient}. We use the cosine $\bar\alpha_t$ noise schedule with offset $s=0.008$ \cite{nichol2021improved}, and we draw timesteps from a \emph{squared-cosine} importance distribution aligned with this schedule. For qualitative sampling and evaluation, we use DDIM with 100 steps, $\eta=0$ (deterministic) \cite{song2020denoising}. Mixed precision training runs on a single node with 4$\times$A100-40GB using \texttt{accelerate} and DDP \cite{vonplaten2022diffusers}.

\renewcommand{\arraystretch}{1.05}
\begin{table}[ht!]
\centering
\caption{Training setup (Diffusers).}
\label{tab:train-setup}
\begin{tabular}{lp{9.5cm}}
\toprule
\textbf{Config} & \textbf{Value} \\
\midrule
Optimizer & AdamW; $\beta_1=0.9$, $\beta_2=0.999$, decay $1\mathrm{e}{-4}$; grad. clip $=1.0$. \\
Base learning rate & $5\mathrm{e}{-4}$; cosine decay with $5\%$ warmup. \\
Batch size per GPU & $64$ (global $256$ over 4 GPUs).  \\
Noise schedule & cosine $\bar\alpha_t$ with offset $s=0.008$ \cite{nichol2021improved}. \\
$t$-sampling distribution & squared-cosine importance over timesteps (in line with the schedule). \\
Loss weighting & MinSNR-$\gamma$ with $\gamma=5$ \cite{hang2023efficient}. \\
Sampler (eval) & DDIM, $100$ steps, $\eta=0$ \cite{song2020denoising}. \\
Epochs & $250$. \\
\bottomrule
\end{tabular}
\end{table}

\paragraph{Compute and Runtime.}
Training uses 4$\times$A100-40GB on a single node with fp16; peak memory per process is $\sim$38\,GB and the 250-epoch run took 15 hours in total.

\subsection{Hyperparameters for Probes and Clustering}\label{probclust}

We list the exact configurations used for linear probing and $k$-means clustering; all hyperparameters and implementation details for linear probe are given in Table~\ref{tab:linprobe}.

\renewcommand{\arraystretch}{1.1}

\begin{table}[h]
\centering
\caption{Linear probe hyperparameters.}
\label{tab:linprobe}
\begin{tabular}{lp{9.5cm}}
\toprule
\textbf{Config} & \textbf{Value} \\
\midrule
Classifier & Linear softmax classifier, PyTorch. \\
Features & GAP-pooled decoder activations at $(t,\ell)$; $\ell_2$-normalized. \\

Loss & Cross-entropy, label smoothing $=0.0$. \\
Optimizer & Adam; lr $=1\mathrm{e}{-3}$; betas $=(0.9,0.999)$; weight decay $=5\mathrm{e}{-4}$. \\
Epochs & 10. \\
Batch size & 512. \\
Regularization & None. \\
\bottomrule
\end{tabular}
\end{table}

\noindent For $k$-means clustering and retrieval (using features at $(t^\star,\ell^\star)$ on the test split), we set $k=70$. The algorithm uses \texttt{n\_init} $=5$ restarts and a maximum of $300$ iterations per run, with a convergence tolerance of $1\times10^{-4}$. When class labels are required, clusters are mapped to classes by majority vote.

\section{Extended Results}\label{app:ext}
This section provides additional experimental results that complement those presented in the main paper. The goal is completeness and clarity, without revisiting interpretations already covered in the main text. 

\subsection{Linear Probe}
The linear-probe results across all decoder readout locations and timesteps are reported in Tables~\ref{ltab:linear_probe_balanced} and~\ref{ltab:linear_probe_unbalanced}, respectively. We report timesteps up to $t=100$ since our sweep indicates that performance peaks at low noise (e.g., $t^\star=25$) and drops sharply at higher noise levels; including larger $t$ values provides little additional insight while increasing feature-extraction cost.

\addtolength{\tabcolsep}{0pt}    
\renewcommand{\arraystretch}{1}

\begin{table}[ht!]
\centering
\footnotesize
\caption{Linear probe test results across decoder stages and denoising timesteps $t$. Each cell reports \emph{Accuracy / Macro F1}. Rows list the three \textit{residual blocks} (\textit{RB1} to \textit{RB3}) at each resolution; attention is used at $16^2$. For each timestep $t$, the best entry appears in \textbf{black}. The selected operating point $(t^{\star}, \ell^{\star})$, where $\ell$ denotes the readout location (resolution $\times$ residual block), is highlighted in \textcolor{darkred}{\textbf{red}}. \textit{Balanced dataset}.}\vspace{1mm}

\begin{tabular}{l c c c c c c c}
\toprule
\multirow{2}{*}{Resolution} & \multirow{2}{*}{\shortstack{Residual\\Block}} & \multicolumn{6}{c}{Denoising timestep $t$ (Accuracy $\uparrow$/ Macro F1 $\uparrow$)} \\
\cmidrule(lr){3-8}
 & & 1 & 10 & 25 & 50 & 75 & 100 \\
\midrule

\multirow{3}{*}{$16^2$ (attn)} 
 & \textit{RB1} & 0.8955 / 0.8822 & 0.8925 / 0.8793 & 0.8890 / 0.8780 & 0.8829 / 0.8668 & 0.8945 / 0.8821 & 0.8910 / 0.8780 \\
 & \textit{RB2} & 0.9133 / 0.9062 & 0.9058 / 0.9021 & 0.9071 / 0.9022 & 0.9090 / 0.9001 & 0.9086 / 0.8974 & 0.9089 / 0.9004 \\
 & \textit{RB3} & \textbf{0.9213} / \textbf{0.9101} & \textbf{0.9191} / \textbf{0.9137} & \textcolor{darkred}{\textbf{0.9240}} / \textbf{0.9114} & \textbf{0.9163} / \textbf{0.9112} & \textbf{0.9159} / \textbf{0.9086} & \textbf{0.9169} / \textbf{0.9098} \\
\midrule
\multirow{3}{*}{$32^2$} 
 & \textit{RB1} & 0.9105 / 0.9044 & 0.9082 / 0.8942 & 0.9076 / 0.8993 & 0.9181 / 0.9087 & 0.9079 / 0.9030 & 0.9057 / 0.9003 \\
 & \textit{RB2} & 0.9019 / 0.8915 & 0.8991 / 0.8900 & 0.9138 / 0.8881 & 0.9157 / 0.9079 & 0.9132 / 0.9011 & 0.8987 / 0.8946 \\
 & \textit{RB3} & 0.9017 / 0.8925 & 0.8971 / 0.8801 & 0.9114 / 0.9050 & 0.9098 / 0.8964 & 0.9146 / 0.9067 & 0.9099 / 0.9033 \\
\midrule
\multirow{3}{*}{$64^2$} 
 & \textit{RB1} & 0.8607 / 0.8464 & 0.8599 / 0.8612 & 0.8574 / 0.8670 & 0.8703 / 0.8629 & 0.8780 / 0.8691 & 0.8763 / 0.8655 \\
 & \textit{RB2} & 0.8079 / 0.7908 & 0.8090 / 0.7880 & 0.8169 / 0.7893 & 0.8307 / 0.8081 & 0.8314 / 0.8112 & 0.8268 / 0.8092 \\
 & \textit{RB3} & 0.7495 / 0.7189 & 0.7512 / 0.7321 & 0.7212 / 0.7478 & 0.7806 / 0.7600 & 0.7895 / 0.7641 & 0.7898 / 0.7659 \\
\midrule
\multirow{3}{*}{$128^2$} 
 & \textit{RB1} & 0.5924 / 0.5428 & 0.5919 / 0.5492 & 0.5937 / 0.5432 & 0.6115 / 0.5725 & 0.6014 / 0.5462 & 0.5774 / 0.5241 \\
 & \textit{RB2} & 0.5513 / 0.4887 & 0.5243 / 0.4570 & 0.5095 / 0.4442 & 0.4807 / 0.3952 & 0.4522 / 0.3802 & 0.4464 / 0.3831 \\
 & \textit{RB3} & 0.4978 / 0.4162 & 0.4380 / 0.3790 & 0.4144 / 0.3342 & 0.3412 / 0.2752 & 0.3228 / 0.2542 & 0.3013 / 0.2350 \\
\bottomrule
\end{tabular}\label{ltab:linear_probe_balanced}
\end{table}

\renewcommand{\arraystretch}{1}
\begin{table}[ht!]
\centering
\footnotesize
\caption{Linear probe test results across decoder stages and denoising timesteps $t$. Each cell reports \emph{Accuracy / Macro F1}. Rows list the three \textit{residual blocks} (\textit{RB1} to \textit{RB3}) at each resolution; attention is used at $16^2$. For each timestep $t$, the best entry appears in \textbf{black}. The selected operating point $(t^{\star}, \ell^{\star})$, where $\ell$ denotes the readout location (resolution $\times$ residual block), is highlighted in \textcolor{darkred}{\textbf{red}}. \textit{Unbalanced dataset}.}\vspace{1mm}

\begin{tabular}{l c c c c c c c}
\toprule
\multirow{2}{*}{Resolution} & \multirow{2}{*}{\shortstack{Residual\\Block}} & \multicolumn{6}{c}{Denoising timestep $t$ (Accuracy $\uparrow$/ Macro F1 $\uparrow$)} \\
\cmidrule(lr){3-8}
 & & 1 & 10 & 25 & 50 & 75 & 100 \\
\midrule

\multirow{3}{*}{$16^2$ (attn)} 
 & \textit{RB1} & 0.8535 / 0.8453 & 0.8512 / 0.8474 & 0.8582 / 0.8488 & 0.8508 / 0.8434 & 0.8582 / 0.8499 & 0.8526 / 0.8426 \\
 & \textit{RB2} & 0.8873 / 0.8798 & 0.8877 / 0.8802 & 0.8886 / 0.8813 & 0.8833 / 0.8785 & 0.8797 / 0.8723 & 0.8778 / 0.8702 \\
 & \textit{RB3} & \textbf{0.9052} / \textbf{0.8974} & \textbf{0.9034} / \textbf{0.8992} & \textcolor{darkred}{\textbf{0.9145}} / {\textbf{0.9064}} & \textbf{0.8998} / \textbf{0.8931} & \textbf{0.8938} / \textbf{0.8873} & \textbf{0.8899} / \textbf{0.8822} \\
\midrule
\multirow{3}{*}{$32^2$} 
 & \textit{RB1} & 0.8965 / 0.8903 & 0.8961 / 0.8892 & 0.8957 / 0.8887 & 0.8975 / 0.8913 & 0.8958 / 0.8898 & 0.8930 / 0.8875 \\
 & \textit{RB2} & 0.8737 / 0.8641 & 0.8739 / 0.8942 & 0.8819 / 0.8728 & 0.8871 / 0.8777 & 0.8900 / 0.8795 & 0.8821 / 0.8728 \\
 & \textit{RB3} & 0.8687 / 0.8597 & 0.8710 / 0.8629 & 0.8777 / 0.8697 & 0.8819 / 0.8751 & 0.8864 / 0.8788 & 0.8803 / 0.8730 \\
\midrule
\multirow{3}{*}{$64^2$} 
 & \textit{RB1} & 0.8276 / 0.8156 & 0.8312 / 0.8291 & 0.8477 / 0.8378 & 0.8400 / 0.8307 & 0.8477 / 0.8396 & 0.8543 / 0.8446 \\
 & \textit{RB2} & 0.7633 / 0.7512 & 0.7713 / 0.7560 & 0.7775 / 0.7594 & 0.7812 / 0.7629 & 0.7773 / 0.7603 & 0.7861 / 0.7733 \\
 & \textit{RB3} & 0.7219 / 0.7043 & 0.7389 / 0.7230 & 0.7525 / 0.7332 & 0.7483 / 0.7286 & 0.7532 / 0.7329 & 0.7615 / 0.7454 \\
\midrule
\multirow{3}{*}{$128^2$} 
 & \textit{RB1} & 0.5601 / 0.5166 & 0.5518 / 0.5102 & 0.5499 / 0.5052 & 0.5613 / 0.5191 & 0.5564 / 0.5103 & 0.5586 / 0.5138 \\
 & \textit{RB2} & 0.4760 / 0.4235 & 0.4732 / 0.4199 & 0.4715 / 0.4192 & 0.4533 / 0.3994 & 0.4568 / 0.4016 & 0.4537 / 0.3997 \\
 & \textit{RB3} & 0.3603 / 0.2923 & 0.3333 / 0.2701 & 0.2924 / 0.2219 & 0.2449 / 0.1834 & 0.2159 / 0.1589 & 0.2152 / 0.1523 \\
\bottomrule
\end{tabular}
\label{ltab:linear_probe_unbalanced}
\end{table}

\subsection{Clustering}
Table~\ref{tab:clustering-summary} reports clustering performance for the U-Net and CNN runs. These metrics are not the result of hyperparameter tuning; rather, they provide a preliminary assessment of representation quality, indicating whether unsupervised clustering is likely to succeed and whether the embedding geometry exhibits class-consistent cohesion and separation. \textit{As the CNN runs are trained with labeled data}, stratified by the true plankton classes, they obtain stronger clustering metrics. By contrast, the U-Net results demonstrate the promising direction of diffusion models for fine-grained representation learning, consistent with the patterns observed in the clustering figure presented in the main article.

\renewcommand{\arraystretch}{1}
\vspace{-3mm}
\begin{table}[h]
\centering
\caption{Clustering quality at the selected $(t^\star,\ell^\star)$ with $k=70$ clusters. Arrows indicate higher is better. \textit{NMI} is mutual information normalized to $[0,1]$; \textit{ARI} is agreement corrected for chance; \textit{Purity} is the fraction assigned to the majority true class within each cluster; \textit{V-measure} is the harmonic mean of homogeneity and completeness; \textit{Silhouette} is the mean coefficient in $[-1,1]$ measuring cohesion vs.\ separation. Results averaged over 5 $k$-means runs.}\vspace{1mm}
\label{tab:clustering-summary}
\setlength{\tabcolsep}{8pt}
\begin{tabular}{l l ccccc}
\toprule
\textbf{Regime} & \textbf{Run} & \textbf{NMI}~$\uparrow$ & \textbf{ARI}~$\uparrow$ & \textbf{Purity}~$\uparrow$ & \textbf{V-measure}~$\uparrow$ & \textbf{Silhouette}~$\uparrow$ \\
\midrule
\multirow{3}{*}{Balanced} 
 & U-Net                 & 0.8688 & 0.7354 & 0.8401 & 0.8642 & 0.2766 \\
 & ResNet~50 (CNN)               & 0.8802 & 0.7918 & 0.8852 & 0.8801 & 0.3067 \\
 & EfficientNet~B0 (CNN)         & 0.9288 & 0.9005 & 0.9219 & 0.9188 & 0.4334 \\
\midrule
\multirow{3}{*}{Unbalanced} 
 & U-Net                 & 0.8450 & 0.7120 & 0.7950 & 0.8500 & 0.2750 \\
 & ResNet~50 (CNN)               & 0.8675 & 0.7629 & 0.8721 & 0.8671 & 0.2932 \\
 & EfficientNet~B0 (CNN)         & 0.9100 & 0.8950 & 0.9180 & 0.9080 & 0.4150 \\
\bottomrule
\end{tabular}
\vspace{-0.5em}
\end{table}

\subsection{Generation Quality and Overfitting}
\begin{figure}[ht!]
    \centering
    \includegraphics[width=0.62\linewidth]{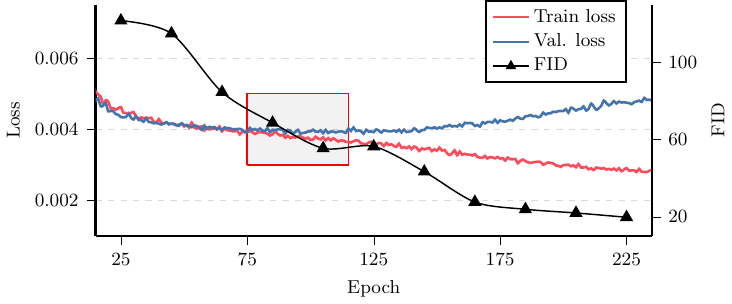}
    \caption{Training and validation loss (left axis) and Fréchet Inception Distance (FID, right axis) on the validation set as a function of epoch for the diffusion model. FID is computed from 10k generated samples compared to the validation set, with grayscale ROIs replicated to three channels for Inception-V3.
}
    \label{fig:overfit_and_fid}\vspace{-3mm}
\end{figure}

Figure~\ref{fig:overfit_and_fid} illustrates how the diffusion model starts to overfit as training progresses. While the training loss (red) continues to decrease monotonically, the validation loss (blue) reaches a minimum and then slowly increases, indicating a loss of generalization. In contrast, the FID score (black, right axis) keeps improving throughout training, meaning that the visual quality of generated samples becomes better even as the model begins to overfit the training distribution, since train and validation distributions differ only mildly due to low intra-class variability. The boxed region around epoch 100 marks the regime where this discrepancy is most evident: the FID is already substantially improved compared to early epochs, but the validation loss has stopped decreasing and begins to drift upward, and further training no longer translates into better performance of the linear probe on downstream classification. For this reason, we selected an early-stopping point at 100 epochs, which balances generation quality and representation quality rather than optimizing FID alone.

\subsection{Motivation Behind MinSNR-$\gamma$ Loss Function}

Training diffusion models with the standard MSE loss induces a strong bias toward low-noise timesteps, where the model is almost reconstructing clean images. This effect originates from the interaction between the noise schedule and the objective: timesteps with a high signal-to-noise ratio automatically receive larger effective weights in the loss, and the corresponding gradients are larger whenever the model makes an error on nearly clean inputs. As a result, these timesteps dominate the optimization, and the network is primarily driven to refine tiny pixel-level corrections around already clean images.

This optimization bias has important consequences for representation learning. When supervision is concentrated on almost clean inputs, the model can rely on very fine, instance-specific details: background texture, small sensor artifacts, or subtle color variations that happen to correlate with the label in the training set. Such cues are easy to exploit at low noise, but they are not stable under moderate corruption or distribution shift. The learned embeddings, therefore, tend to encode brittle, image-specific information rather than robust, class-level structure, which limits generalization in fine-grained classification, especially when images are perturbed.

In contrast, the proposed MinSNR-$\gamma$ loss with $\gamma=5$ explicitly counteracts this bias by reshaping the learning signal across timesteps. Instead of allowing high-SNR (low-noise) inputs to dominate the gradients, it down-weights very clean timesteps and amplifies the contribution of a band of mid-noise timesteps, where the input is still semantically recognizable but low-level pixel cues have largely been suppressed. To perform well in this regime, the model must rely on features that summarize stable, class-relevant structure (e.g., the shape and configuration of object parts) rather than idiosyncratic texture or background artifacts. Empirically, this reweighting leads to more robust representations: in our experiments, MinSNR-$\gamma$ preserves downstream classification performance under moderate corruption, whereas models trained with the standard MSE objective exhibit a marked degradation in that regime.

\vspace{2mm}

\begin{figure}[ht!]
    \centering
    \includegraphics[width=0.6\linewidth]{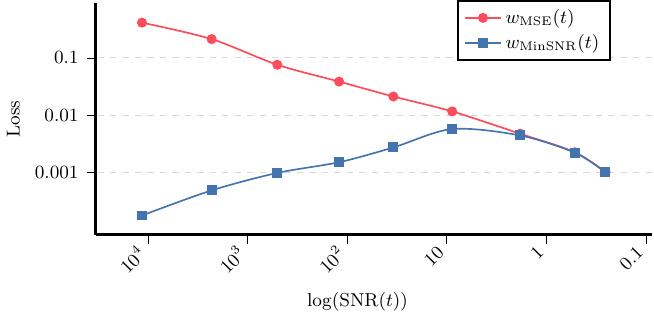}
\caption{Mean-squared denoising loss of the U-Net as a function of the signal-to-noise ratio (SNR). The loss remains low at high SNR and increases sharply as $\gamma$ decreases, indicating that the model struggles to reconstruct highly noisy inputs.}

    \label{fig:snr-loss}
\end{figure}

Figure~\ref{fig:snr-loss} illustrates how this mechanism manifests in practice by plotting the per-timestep denoising loss of a U-Net as a function of the signal-to-noise ratio. Under the standard MSE objective (blue curve), the loss is largest for very high SNR values, confirming that optimization is dominated by nearly clean inputs. In contrast, the MinSNR-$\gamma$ objective (red curve) strongly suppresses the loss in this high-SNR regime and instead concentrates the effective training signal in an intermediate band of SNR values, where the inputs are noisy yet still semantically informative. Even if the normalized per-timestep loss curves appear superficially similar, the aggregate objective integrates them with different timestep-dependent weights, so that MinSNR-$\gamma$ yields a more balanced training signal whose gradients are better aligned with the requirements of downstream classification across a range of noise levels.

\end{document}